# Solving a Machine Learning Regression Problem Based on the Theory of Random Functions

Bakhvalov Y. N., Ph.D., Independent Researcher, Cherepovets, Russia

**Abstract.**

This paper studies a machine learning regression problem as a multivariate approximation problem using the framework of the theory of random functions. An ab initio derivation of a regression method is proposed, starting from postulates of indifference. It is shown that if a probability measure on an infinite-dimensional function space possesses natural symmetries (invariance under translation, rotation, scaling, and Gaussianity), then the entire solution scheme – including the kernel form, the type of regularization, and the noise parameterization – follows analytically from these postulates. The resulting kernel coincides with a generalized polyharmonic spline; however, unlike existing approaches, it is not chosen empirically but arises as a consequence of the indifference principle. This result provides a theoretical foundation for a broad class of smoothing and interpolation methods, demonstrating their optimality in the absence of a priori information.

A previous version of this work was published by the author as [20], itself a development of ideas from [1]. The present paper substantially expands and revises the original material: the problem is reformulated as an approximation problem, a rigorous description via a probability measure is introduced, mathematical inaccuracies are corrected, and new transformations and justifications are added. The mathematical apparatus and the structure of the exposition have been completely reworked.

Let the training set consist of input vectors $x_1, x_2, \dots, x_k (x_i \in R^n)$ and output values $y_1, y_2, \dots, y_k$ $(y_i \in R)$.

The regression problem can then be cast as an approximation problem of the form

$$y_i = f(x_i) + u_i \tag{1}$$

where $f(x)$ – is the function linking inputs $x_i$ to outputs $y_i$ – the target model to be inferred from the training data;

and $u_1, u_2, \dots, u_k \in R$ - are independent random variables, normally distributed with zero mean, which can be regarded as having been added to $f(x_i)$, so that $f(x_i)$ may not exactly match $y_i$ (thereby modeling errors or ambiguities present in the training sample).

The problem may also be interpreted as follows.

Suppose there existed some unknown function $f(x)$. Experiments were then performed on it: for a sample $x_1, x_2, \dots, x_k$ the values $f(x_i)$ were obtained (but are unknown to us), to which random, unknown quantities $u_1, u_2, \dots, u_k$ were added. As a result, in addition to the known $x_1, x_2, \dots, x_k$ we obtained $y_1, y_2, \dots, y_k$ . The task now is, given the sequences $x_i$ and $y_i$ to infer what $f(x)$ could have been.

At first glance (even if the characteristics of $u_i$ are known), this setup may seem unhelpful, since nothing is known about the nature of $f(x)$, and equation (1) places no restrictions on what $f(x)$ might be.

Let us first consider a strongly simplified version of the problem.

Assume we have a "hint." Suppose someone has provided us in advance with possible correct answers-candidate functions $f_j(x)$ (a finite list) – along with prior probabilities $p_j$ that each is the correct one. However, these probabilities are prior in the sense that they were specified without knowledge of the subsequent experiments on $f(x)$ and the additional independent information we now have in the form of $x_i$ and $y_i$.

Assume the variances of the random variables $u_1, u_2, \dots, u_k$ are known and identical, equal to $\sigma^2$. Then the probabilities that the j-th candidate $f_j(x)$ is the correct answer are proportional to

$$\widehat{p_j} = p_j (2\pi\sigma^2)^{-k/2} exp\left(-\frac{1}{2\sigma^2}\sum_{i=1}^{k} u_i^2\right), \qquad (2)$$

where

$$u_i = y_i - f_j(x_i)$$

By evaluating all $\widehat{p_j}$ and selecting the maximum, we pick the best $f_j(x)$ as the best solution to the approximation problem (1).

But what is a set of possible functions $f(x)$ endowed with probabilities (i.e., a probability distribution over functions)? This is precisely a description of a random function.

As seen above, given such a hint (a random function with a finite set of realizations), we immediately obtain a solution. Analogous reasoning, however, can be carried out over a much larger set of functions and without an explicit "hint".

Take as the candidate set $\mathcal{F}$ the set of all continuous real-valued functions on $R^n$. This set can be viewed in an infinite-dimensional Hilbert space, with each "dimension" corresponding to the value of the function at some particular $x \in R^n$.

On this set, one can in turn define a probability measure $\mu$ (which together with $\mathcal{F}$ specifies a random function).

It is worth noting that on this set-being an uncountable infinite-dimensional space (in which a Lebesgue measure does not exist) - a classical probability density is not defined. Nevertheless, one can specify a probability measure $\mu$ as the projective limit of consistent finite-dimensional distributions.

The correctness of this approach is standard in the theory of Gaussian measures on infinite-dimensional spaces (Bogachev, 1998 [3]; Vakhania et al., 1987 [16]). The existence of $\mu$ under given consistent finite-dimensional distributions is ensured by Kolmogorov's extension theorem (Kallenberg, 2002 [11]; Gikhman & Skorokhod, 1971 [9]).

We treat the probability measure $\mu$ as unknown. However, guided by the principle of indifference, we may posit that μ satisfies certain symmetries (which could reasonably have been assumed before any experiments on $f(x)$ were performed).

1. Invariance of μ under translations and rotations of the coordinate system.

For any rotation matrix $M$, any translation vector $t$, and the mapping $T: f_1 \to f_2$ defined by

$$f_2(x) = f_1(Mx + t), x, t \in R^n \tag{3}$$

we require that for every measurable subset $A \subset \mathcal{F}$ one has $\mu(A) = \mu\big(T(A)\big)$.

That is, any two subsets of functions that are mapped into one another by a rotation or a translation must have the same probability under $\mu$.

2. Invariance of μ under scaling transformations.

For any $k \in R$ and the mapping $T: f_1 \to f_2$ defined by

$$f_2(x) = kf_1(x/k), x \in R^n \tag{4}$$

we require that for every measurable subset $A \subset \mathcal{F}$ one has $\mu(A) = \mu\big(T(A)\big)$.

Thus, any two subsets of functions related by a change of scale must have the same probability under $\mu$.

3. The measure $\mu$ is an infinite-dimensional Gaussian measure with zero mean.

This means that any finite-dimensional projection of the random function is multivariate normal, and the measure is Kolmogorov – consistent (Gikhman & Skorokhod [9]).

To clarify item 3, we state two corollaries with an intuitive interpretation.

Corollary 3.1. From Gaussianity of $\mu$ it follows that if a subset of functions is generated by scaling a single base function $f_1(x) \to kf_1(x)$, then the conditional probability density on this subset is Gaussian in the parameter k with zero mean.

That is, if we assume the solution lies within the family

$$f_k(x) = kf_1(x), k \in R \tag{5}$$

where $k$ is a real scalar (the "amplitude" of $f_1(x)$) and $f_1(x)$ is any fixed continuous function used to generate the set (5).

Then the conditional distribution over this set is normal in the amplitude k with mean zero.

In other words, among two realizations that differ only by amplitude, the one with smaller amplitude is more probable; the distribution in $k$ is Gaussian with zero mean. We do not specify the variance (assumed nonzero); the point here is the Gaussian character of the law.

Corollary 3.2. If we add an arbitrary fixed function$f_2(x)$ to each function in the set from Corollary 3.1, the distribution over k remains Gaussian (although the mean may shift).

Adding the same $f_2(x)$ to all functions in that family yields a new family

$$f_k^*(x) = kf_1(x) + f_2(x), k \in R \tag{6}$$

with $f_1(x)$ and $f_2(x)$ any continuous functions.

Regardless of the choice of $f_2(x)$, the probability density on this subset (as a finite-dimensional projection induced by $\mu$) remains normal in the amplitude $k$, though the mean need not be zero.

Intuitively, if we take an arbitrary "baseline" function $f_2(x)$ and then add to it scaled versions of some continuous pattern $f_1(x)$, the amplitudes $k$ of these oscillations around $f_2(x)$ are distributed normally (with a possibly shifted mean).

The choice of symmetries in items 1-3 for the probability measure μ is motivated by the principle of indifference in the complete absence of prior information before observing the training data. In this situation it is most natural to assume:

1. **Spatial invariance (1):** $\mu$ should not depend on the choice of the origin (translation $t$) or the orientation (rotation $M$). There is no distinguished point or direction in $R^n$.

2. **Scale invariance (2):** There is no preferred observation scale. The laws linking x and y should retain the same form under changes of units (scaling by $k$).

3. **No privileged amplitudes (3):** Within any given “pattern” $f_1(x)$, there is no a priori preferred amplitude of its manifestation. The most general and natural assumption for the distribution of amplitudes is the normal law (Gaussianity), which follows from the maximum entropy principle under a fixed variance.

Thus, postulates 1-3 constitute a minimal set of natural assumptions about symmetries of the function space in the absence of any other prior information.

Let us now consider the consequences of items 1-3 for a random function satisfying them.

Any nondegenerate (as in our case) multivariate normal distribution can be reduced to a collection of independent normal random variables. This implies that there exists a sequence of functions (forming a basis with respect to which any continuous function can be expanded) such that the random function under consideration can be written as a linear combination of that sequence with independent normal coefficients. This is precisely the canonical expansion of a random function; hence such a random function must admit a canonical expansion.

The canonical expansion of a random function (V. S. Pugachev [13], pp. 248-249) for a countable basis is

$$F(x) = m_f(x) + \sum_{j=1}^{\infty} V_j \varphi_j(x), \tag{7}$$

where $m_f(x)$ is the mean function;

$V_j$ – are uncorrelated random variables with zero means (the coefficients of the canonical expansion);

$\varphi_j(x)$ – are the coordinate functions of the expansion.

From item 3 we have $m_f(x) = 0$. Moreover, since in our case the family $\varphi_j(x)$ is uncountable, (7) generalizes to an integral canonical representation:

$$F(x) = \int_{\Lambda} V(\lambda)\varphi(x,\lambda)d\lambda, \tag{8}$$

where $V(\lambda)$ is white noise indexed by $\lambda$,

and $\varphi(x,\lambda)$ is a (nonrandom) function of $x$ and the parameter $\lambda$.

In this way, every realization (a continuous function) is expanded in terms of uncorrelated infinitesimal elementary random functions $V(\lambda)\varphi(x,\lambda)d\lambda$. At this stage, however, we have only established from item 3 (Gaussianity) the existence of such a representation, not its specific form.

Let us now examine the existence of the canonical integral representation (8), which follows from item 3, in light of item 1. From invariance of the measure under all translations and rotations applied to any subsets of realizations, it follows that the random function is strictly stationary and isotropic. Hence it is not only representable in the form (8), but it must also admit a spectral representation.

The integral canonical representation of a stationary random function is given by V. S. Pugachev [13, p. 333]. The white noise term used there, whose intensity may vary across frequencies, can be rewritten in the form

$$V(\omega) \to \frac{1}{2} V(\omega) \sqrt{\frac{S(\omega)}{d\omega}}\,, \tag{9}$$

where $V(\omega)$ is now taken to have the same unit intensity (variance) at all frequencies (the correctness of using (9) will be further justified below).

In general, (9) is a formal change of representation for white noise $V(\omega)$ that enforces a constant unit intensity across frequencies. This is standard in the spectral theory of random processes (Cramér, 1942 [7]) and is equivalent to the more rigorous formulation via orthogonal random measures. The factor 1/2 accounts for restricting the integration domain to $R_1^n$ and enforcing the reality condition (11).

Thus, we can write our random function as

$$F(x) = \frac{1}{2} \int_{R^n} V(\omega) \sqrt{\frac{S(\omega)}{d\omega}} e^{i\omega x} d\omega\,, \quad x, \omega \in R^n \tag{10}$$

where $S(\omega)$ is the spectral density (a nonnegative real even function), and $V(\omega)$ is a complex-valued random function such that each value associated with an infinitesimal frequency element $d\omega$ is an independent random variable (subject to an additional condition below), with real and imaginary parts distributed normally with zero mean and unit variance. That is, $V(\omega)$ is white noise with the same variance at all frequencies.

Additional condition. Since all realizations $f(x)$ must be real-valued, we require $V(\omega)$ and $V(-\omega)$ to be complex conjugates for all $\omega$.

To encode this, split the frequency space $R^n$ into two parts, $R_1^n$ and $R_2^n$, by any hyperplane of dimension $n-1$ passing through the origin. Then we can write

$$V(\omega)=\begin{cases}V_R(\omega)+iV_I(\omega), if\ \omega\in R_1^n\\ V_R(\omega)-iV_I(\omega), if\ \omega\in R_2^n\end{cases}, \tag{11}$$

where $V_R(\omega)$ and $V_I(\omega)$ are real, symmetric random functions (i.e., $V_R(\omega)=V_R(-\omega)$, $V_I(\omega)=V_I(-\omega)$). Each value associated with an element $d\omega$ (as in (10)) is an independent random variable with zero mean and unit variance (real symmetric white noise).

Using this, (10) can be rewritten as

$$F(x)=\frac{1}{2}\int_{R_1^n}\big(V_R(\omega)+iV_I(\omega)\big)\sqrt{\frac{S(\omega)}{d\omega}}e^{i\omega x}d\omega+$$

$$+\frac{1}{2}\int_{R_2^n}\big(V_R(\omega)-iV_I(\omega)\big)\sqrt{\frac{S(\omega)}{d\omega}}e^{i\omega x}d\omega \tag{12}$$

Since for any $\omega$, if $\omega\in R_1^n$ then $-\omega\in R_2^n$ (and vice versa), in the second integral we may change variables to integrate over $R_1^n$ and simultaneously replace $\omega$ by $-\omega$ in the integrand. This yields

$$F(x)=\frac{1}{2}\int_{R_1^n}\big(V_R(\omega)+iV_I(\omega)\big)\sqrt{\frac{S(\omega)}{d\omega}}e^{i\omega x}d\omega$$

$$+\frac{1}{2}\int_{R_1^n}\big(V_R(-\omega)-iV_I(-\omega)\big)\sqrt{\frac{S(-\omega)}{d\omega}}e^{-i\omega x}d\omega=$$

$$=\frac{1}{2}\int_{R_1^n}\sqrt{\frac{S(\omega)}{d\omega}}\Big(V_R(\omega)\big(e^{i\omega x}+e^{-i\omega x}\big)+iV_I(\omega)\big(e^{i\omega x}-e^{-i\omega x}\big)\Big)d\omega=$$

$$=\int_{R_1^n}\sqrt{\frac{S(\omega)}{d\omega}}\left(V_R(\omega)\cos(\omega x)-V_I(\omega)\sin(\omega x)\right)d\omega \tag{13}$$

Thus we obtain an integral canonical representation of our random function (corresponding to (8)), expressed in terms of real functions and independent real normal random variables with unit variance.

As the transformations show, expression (13) is fully equivalent to the complex form (10), in which white noise is represented as

$$\frac{1}{2}V(\omega)\sqrt{\frac{S(\omega)}{d\omega}}$$

For convenience in the subsequent derivations, we replace the integral in (13) by an infinite sum. Let $v_j^R$ denote the infinite sequence of values of $V_R(\omega)$, each corresponding to its own small integration cell d$\omega$ within $R_1^n$ (with associated frequency $\omega_j$). Define analogously $v_j^I$ for $V_I(\omega)$, and $s_j$ for $S(\omega)$. Then (13) can be written as

$$F(x) = \sum_{j=0}^{\infty} \sqrt{\frac{s_j}{d\omega}} \left(v_j^R \cos(\omega_j x) - v_j^I \sin(\omega_j x)\right) d\omega \tag{14}$$

With the understanding that $d\omega$ denotes a very small cell (in the limit, infinitesimal), so that (14) tends to (13) as $d\omega \to 0$.

Equivalently, we can rewrite (14) as

$$F(x) = \sum_{j=0}^{\infty} v_j^R \sqrt{s_j d\omega} \cos(\omega_j x) - \sum_{j=0}^{\infty} v_j^I \sqrt{s_j d\omega} \sin(\omega_j x) \tag{15}$$

Expression (15) is precisely the canonical expansion of the form (7) with zero mean, where $v_j^R$ and $v_j^I$ are independent random variables with unit variance, and the coordinate functions are $\sqrt{s_j d\omega}\cos(\omega_j x)$ and $-\sqrt{s_j d\omega}\sin(\omega_j x)$.

Using the canonical expansion via these coordinate functions, the two-point correlation function is

$$K_f(x_1, x_2) = \sum_{j=0}^{\infty} s_j \left(\cos(\omega_j x_1)\cos(\omega_j x_2) + \sin(\omega_j x_1)\sin(\omega_j x_2)\right) d\omega$$
$$= \sum_{j=0}^{\infty} s_j \cos\left(\omega_j (x_1 - x_2)\right) d\omega \tag{16}$$

i.e., we obtain the autocorrelation function (as expected for a stationary random function):

$$k_f(\tau) = \sum_{j=0}^{\infty} s_j \cos(\omega_j \tau)\, d\omega \tag{17}$$

Returning from (17) to the integral form gives

$$k_f(\tau) = \int_{R_1^n} S(\omega)\cos(\omega\tau)\,d\omega \quad (18)$$

Which is the standard canonical representation of the correlation function, with $S(\omega)$ the spectral density.

Thus we have shown that representing the white noise in the integral canonical decomposition of a stationary random function via $V(\omega)$ (as in (11)) together with the spectral density $S(\omega)$ can be realized through (9), and the random function itself can be written either in the complex form (10) or the real form (13).

Let us now see how to express the most probable realization of the random function (10) (or equivalently its representations (13) or (14)) that best fits the training sample (the original task).

Since $\mu$ is a Gaussian measure and is generated by independent normal coefficients $v_j^R$ and $v_j^I$ in the canonical expansion (15), it is natural to consider its finite-dimensional projections. For any finite collection of N coefficients $v_j^R$and $v_j^I$, the corresponding probability density is proportional to

$$exp\left(-\frac{1}{2}\sum_{j=0}^{N}\left(\left(v_j^R\right)^2 + \left(v_j^I\right)^2\right)\right) \quad (19)$$

Although in the infinite-dimensional limit $N \to \infty$ this density as a function of all coefficients does not exist in a strict sense, the expression

$$exp\left(-\frac{1}{2}\sum_{j=0}^{\infty}\left(\left(v_j^R\right)^2 + \left(v_j^I\right)^2\right)\right) \quad (20)$$

remains meaningful as a functional that defines the relative “weight” of realizations for a MAP estimate. Maximizing (20) (in light of the data) yields the most probable realization – this is standard when working with Gaussian measures on function spaces (Rasmussen & Williams [14]).

Moreover, from items 1 and 2 (invariance of μ under the transformations (3) and (4)) it follows that for any pair of functions related by (3) or (4), the limiting ratio of such functionals (as one passes from (19) to (20), or further to the integral form) equals one.

Expressions (19) - (20), however, account only for the prior and ignore the $u_i$. Suppose the random variables $u_i$ in (1) are i.i.d. with variance $\sigma^2$. Multiplying (20) by the joint normal likelihood of the $u_i$, the exponential part becomes

$$exp\left(-\frac{1}{2}\sum_{j=0}^{\infty}\left(\left(v_j^R\right)^2+\left(v_j^I\right)^2\right)-\frac{1}{2\sigma^2}\sum_{i=1}^{k}u_i^2\right) \tag{21}$$

Thus (by the same logic as in (2)), finding the best realization in (14) or (15) reduces to finding sequences $v_j^R$ and $v_j^I$ that maximize (21). Equivalently, this is the minimization problem

$$\frac{1}{2}\sum_{j=0}^{\infty}\left(\left(v_j^R\right)^2+\left(v_j^I\right)^2\right)+\frac{1}{2\sigma^2}\sum_{i=1}^{k}u_i^2 \rightarrow \min \tag{22}$$

From (1) and (15) the training sample induces the system

$$\begin{cases}\sum_{j=0}^{\infty}v_j^R\sqrt{s_j d\omega}\cos(\omega_j x_1)-\sum_{j=0}^{\infty}v_j^I\sqrt{s_j d\omega}\sin(\omega_j x_1)+u_1=y_1\\ \sum_{j=0}^{\infty}v_j^R\sqrt{s_j d\omega}\cos(\omega_j x_2)-\sum_{j=0}^{\infty}v_j^I\sqrt{s_j d\omega}\sin(\omega_j x_2)+u_2=y_2\\ \cdots\\ \sum_{j=0}^{\infty}v_j^R\sqrt{s_j d\omega}\cos(\omega_j x_k)-\sum_{j=0}^{\infty}v_j^I\sqrt{s_j d\omega}\sin(\omega_j x_k)+u_k=y_k\end{cases} \tag{23}$$

We thus obtain a minimization problem (22) subject to the equality constraints (23). This is a variational problem in a Hilbert space with a unique solution (for $\sigma > 0$), and it can be solved via the method of Lagrange multipliers.

The Lagrangian takes the form

$$L(v_1^R,v_2^R,..,v_1^I,v_2^I,..,u_1,..,u_k,\lambda_1,..,\lambda_k)=\frac{1}{2}\sum_{j=0}^{\infty}\left(\left(v_j^R\right)^2+\left(v_j^I\right)^2\right)+\frac{1}{2\sigma^2}\sum_{i=1}^{k}u_i^2+$$

$$+\sum_{i=1}^{k}\lambda_i\left(y_i-u_i-\sum_{j=0}^{\infty}v_j^R\sqrt{s_j d\omega}\cos(\omega_j x_i)+\sum_{j=0}^{\infty}v_j^I\sqrt{s_j d\omega}\sin(\omega_j x_i)\right), \tag{24}$$

where $\lambda_i$ are the Lagrange multipliers (one per data point).

From the conditions $\frac{dL}{d\lambda_i}=0$ we recover the constraint system (23).

From $\frac{dL}{dv_j^R}=0$ we obtain

$$v_j^R = \sqrt{s_j d\omega} \sum_{i=1}^{k} \lambda_i \cos(\omega_j x_i) \tag{25}$$

From $\frac{dL}{dv_j^I} = 0$ we obtain

$$v_j^I = -\sqrt{s_j d\omega} \sum_{i=1}^{k} \lambda_i \sin(\omega_j x_i) \tag{26}$$

From $\frac{dL}{du_i} = 0$ we obtain

$$u_i = \sigma^2 \lambda_i \tag{27}$$

Thus, the discrepancies $u_i$ between the best-fit $f(x)$ and the training data in (1) are proportional to the Lagrange multipliers, with proportionality factor $\sigma^2$ (the variance of $u_i$).

Substituting (25) - (26) into (15) gives the most probable realization $f(x)$:

$$f(x) = \sum_{j=0}^{\infty} s_j \sum_{i=1}^{k} \lambda_i \cos(\omega_j x_i) \cos(\omega_j x)\, d\omega + \sum_{j=0}^{\infty} s_j \sum_{i=1}^{k} \lambda_i \sin(\omega_j x_i) \sin(\omega_j x)\, d\omega =$$

$$= \sum_{i=1}^{k} \lambda_i \sum_{j=0}^{\infty} s_j \cos\Big(\omega_j (x_i - x)\Big)\, d\omega \tag{28}$$

The inner sums are exactly the canonical expansions of the correlation function (cf. (16)). Hence

$$f(x) = \sum_{i=1}^{k} \lambda_i k_f(x_i - x) \tag{29}$$

Thus, under assumptions 1 and 3, the solution to (1) is a linear combination of correlation functions.

To determine the coefficients $\lambda_i$ in (29), substitute (25), (26), and (27) into the constraint system (23). This yields

$$\begin{cases} \sum_{i=1}^{k} \lambda_i k_f(x_i - x_1) + \sigma^2 \lambda_1 = y_1 \\ \sum_{i=1}^{k} \lambda_i k_f(x_i - x_2) + \sigma^2 \lambda_2 = y_2 \\ \dots \\ \sum_{i=1}^{k} \lambda_i k_f(x_i - x_k) + \sigma^2 \lambda_k = y_k \end{cases} \tag{30}$$

In matrix form ($\sigma^2$ adds to the main diagonal):

$$(K + \sigma^2 E)\lambda = Y, \tag{31}$$

where $K$ is the $(k \times k)$ matrix with entries $k_{ij} = k_f(x_i - x_j)$,

$E$ is the identity matrix,

$\lambda$ – column vector $(\lambda_1, \lambda_2, \dots, \lambda_k)$,

$Y$ – column vector $(y_1, y_2, \dots, y_k)$.

Therefore,

$$\lambda = (K + \sigma^2 E)^{-1} Y \tag{32}$$

If $\sigma^2 = 0$ in (32), the approximation problem (1) becomes exact interpolation. By tuning $\sigma^2$ one controls the fidelity of $f(x)$ to the training data in (1), helping to avoid overfitting.

At this point, using (29) and (32) yields a solution to (1). However, these derivations involved the spectral density $S(\omega)$ (entering via the correlation function in (18)), which we have not yet specified.

Let $S_{f_1}(\omega)$ denote the spectral representation of a particular realization $f_1(x)$ of the random function (10):

$$S_{f_1}(\omega) = \frac{1}{2} V_1(\omega) \sqrt{\frac{S(\omega)}{d\omega}}, \tag{33}$$

where $V_1(\omega)$ is a particular realization of $V(\omega)$, i.e., the collection of values taken by $V(\omega)$ that turns (10) into the specific realization $f_1(x)$. Then

$$f_1(x) = \int_{R^n} S_{f_1}(\omega) e^{i\omega x} d\omega \tag{34}$$

Similarly, let $S_{f_2}(\omega)$ be the spectral representation of another function $f_2(x)$. Assume $f_1(x)$ and $f_2(x)$ satisfy (4). Then

$$f_2(x) = \int_{R^n} S_{f_2}(\omega) e^{i\omega x} d\omega = k f_1\left(\frac{x}{k}\right) = k \int_{R^n} S_{f_1}(\omega) e^{i\omega \frac{x}{k}} d\omega =$$

$$= \int_{R^n} k^{n+1} S_{f_1}(\omega) e^{i\omega \frac{x}{k}} d\frac{\omega}{k} = \int_{R^n} k^{n+1} S_{f_1}(\omega k) e^{i\omega x} d\omega \qquad (35)$$

Comparing the first and last expressions in (35) yields the relation between the spectral representations of $f_1(x)$ and $f_2(x)$ that satisfy (4):

$$S_{f_2}(\omega) = k^{n+1} S_{f_1}(\omega k) \qquad (36)$$

Returning to (33), rewrite it by solving for $V_1(\omega)$:

$$V_1(\omega) = 2 S_{f_1}(\omega) \sqrt{\frac{d\omega}{S(\omega)}} \qquad (37)$$

Since $f_1(x)$ and $f_2(x)$ obey (4), as noted above, the limiting ratio of their functionals (20) must equal one.

The term under the sum in (20) can be rewritten as

$$\left(v_j^R\right)^2 + \left(v_j^I\right)^2 = \left|V(\omega_j)\right|^2 \qquad (38)$$

For frequency $\omega_j$, the squared modulus of (37) gives

$$\left|V_1(\omega_j)\right|^2 = 4 \frac{\left|S_{f_1}(\omega_j)\right|^2}{S(\omega_j)} d\omega \qquad (39)$$

Substituting (39) into (20) yields

$$exp\left(-2 \sum_{j=0}^{\infty} \frac{\left|S_{f_1}(\omega_j)\right|^2}{S(\omega_j)} d\omega\right) \qquad (40)$$

In the limit, this sum becomes the integral

$$exp\left(-2 \int_{R_1^n} \frac{\left|S_{f_1}(\omega)\right|^2}{S(\omega)} d\omega\right) = exp\left(- \int_{R^n} \frac{\left|S_{f_1}(\omega)\right|^2}{S(\omega)} d\omega\right) \qquad (41)$$

Therefore, for $f_1(x)$ and $f_2(x)$, the fact that they have the same “weight” as realizations – expressed by the unit limiting ratio of their functionals, now written as in (41) – can be enforced by equating the integral parts:

$$\int_{R^n} \frac{\left|S_{f_1}(\omega)\right|^2}{S(\omega)} d\omega = \int_{R^n} \frac{\left|S_{f_2}(\omega)\right|^2}{S(\omega)} d\omega \quad (42)$$

Substituting the previously obtained relation (36) into (42) gives

$$\int_{R^n} \frac{\left|S_{f_1}(\omega)\right|^2}{S(\omega)} d\omega = \int_{R^n} \frac{\left|S_{f_2}(\omega)\right|^2}{S(\omega)} d\omega = \int_{R^n} \frac{k^{2n+2}\left|S_{f_1}(\omega k)\right|^2}{S(\omega)} d\omega =$$

$$= \int_{R^n} \frac{k^{n+2}\left|S_{f_1}(\omega k)\right|^2}{S(\omega)} d(\omega k) = \int_{R^n} \frac{k^{n+2}\left|S_{f_1}(\omega)\right|^2}{S\left(\frac{\omega}{k}\right)} d\omega \quad (43)$$

Comparing the first and last expressions in (43) yields the scaling relation for the spectral density:

$$\frac{S(\omega/k)}{S(\omega)} = k^{n+2} \quad (44)$$

Since the random function under consideration is stationary, the autocorrelation function (18) must be radially symmetric. Consequently, its spectral density must also be radially symmetric.

A function satisfying (44) and radial symmetry is

$$S(\omega) = a\|\omega\|^{-(n+2)}, \quad (45)$$

where a is a constant.

Since $\omega \in R^n$ with components $(\omega_1, \omega_2, \dots, \omega_n)$, (45) can also be written as

$$S(\omega_1, \omega_2, \dots, \omega_n) = a(\omega_1^2 + \omega_2^2 + \cdots + \omega_n^2)^{-\frac{n+2}{2}}, \omega_1, \omega_2, \dots, \omega_n \in R \quad (46)$$

As is clear, (44) holds for any coefficient $a$ in (45) - (46); $a$ may be chosen so that $k_f(0)$ represents the variance of the random function. The ratio between $k_f(0)$ and $\sigma^2$ determines how closely $f(x)$ should fit the training data. As seen from (32) and (29), if we simultaneously multiply $k_f$ and $\sigma^2$ by the same factor, the function (29) remains unchanged.

Let us examine the result (45) - (46) in more detail. Denote $\tau$ in $k_f(\tau)$ from (18) by the vector $(\tau_1, \tau_2, \dots, \tau_n)$. Using the spectral density (46) and passing from the domain $R_1^n$ to $R^n$, we obtain

$$k_f(\tau_1, \tau_2, \dots, \tau_n) = \frac{a}{2} \int_{-\infty}^{\infty} \dots \int_{-\infty}^{\infty} \frac{\cos(\omega_1\tau_1 + \omega_2\tau_2 + \cdots + \omega_n\tau_n)}{(\omega_1^2 + \omega_2^2 + \cdots + \omega_n^2)^{\frac{n+2}{2}}} d\omega_n d\omega_{n-1} \dots d\omega_1 \quad (47)$$

What is the cross-section of the correlation function (47), for example when $\tau_n$=0? Consider the innermost integral in (47) under $\tau_n$=0:

$$\int_{-\infty}^{\infty} \frac{\cos(\omega_1\tau_1 + \omega_2\tau_2 + \cdots + \omega_{n-1}\tau_{n-1})}{(\omega_1^2 + \omega_2^2 + \cdots + \omega_n^2)^{\frac{n+2}{2}}} d\omega_n =$$

$$= 2\cos(\omega_1\tau_1 + \omega_2\tau_2 + \cdots + \omega_{n-1}\tau_{n-1}) \int_0^{\infty} \frac{d\omega_n}{(\omega_1^2 + \omega_2^2 + \cdots + \omega_n^2)^{\frac{n+2}{2}}} \quad (48)$$

Make the substitution

$$\frac{\omega_1^2 + \omega_2^2 + \cdots + \omega_{n-1}^2}{\omega_1^2 + \omega_2^2 + \cdots + \omega_{n-1}^2 + \omega_n^2} = 1 - t \quad (49)$$

After carrying out the transformation, (48) becomes

$$\frac{\cos(\omega_1\tau_1 + \omega_2\tau_2 + \cdots + \omega_{n-1}\tau_{n-1})}{(\omega_1^2 + \omega_2^2 + \cdots + \omega_{n-1}^2)^{\frac{(n-1)+2}{2}}} \int_0^1 t^{-\frac{1}{2}}(1-t)^{\frac{n-1}{2}} dt =$$

$$= \frac{\cos(\omega_1\tau_1 + \omega_2\tau_2 + \cdots + \omega_{n-1}\tau_{n-1})}{(\omega_1^2 + \omega_2^2 + \cdots + \omega_{n-1}^2)^{\frac{(n-1)+2}{2}}} B\left(\frac{1}{2}, \frac{n+1}{2}\right), \quad (50)$$

where B is Euler's beta function.

Therefore, the cross-section of (47) at $\tau_n$=0 is

$$k_f(\tau_1, \dots, \tau_{n-1}) =$$

$$= \frac{a}{2} B\left(\frac{1}{2}, \frac{n+1}{2}\right) \int_{-\infty}^{\infty} \dots \int_{-\infty}^{\infty} \frac{\cos(\omega_1\tau_1 + \cdots + \omega_{n-1}\tau_{n-1})}{(\omega_1^2 + \omega_2^2 + \cdots + \omega_{n-1}^2)^{\frac{(n-1)+2}{2}}} d\omega_{n-1} \dots d\omega_1 \quad (51)$$

Since the factor $\frac{a}{2} B\left(\frac{1}{2}, \frac{n+1}{2}\right)$ plays the role of an overall constant, comparing (51) with (47) shows that taking a hyperplane cross-section through the origin in $R^n$ yields, up to a constant factor, the correlation function in $R^{n-1}$ expressed again in the form (47) (with a correspondingly adjusted coefficient $a$ in (45) - (46)).

Hence, to determine the correlation function it suffices to find it in the one-dimensional case on $\tau \in [0, +\infty)$ and then, using radial symmetry, extend it to spaces of any dimension.

For the one-dimensional case, the spectral density (45) with $a = 1$ is

$$S(\omega) = |\omega|^{-3}, \omega \in R \quad (52)$$

The random function in the form (13) then becomes, in one dimension,

$$F(x) = \int_0^\infty \frac{V_R(\omega)\cos(\omega x) - V_I(\omega)\sin(\omega x)}{\omega\sqrt{\omega d\omega}} d\omega, \qquad x, \omega \in R \tag{53}$$

The corresponding correlation function (18) in one dimension is

$$k_f(\tau) = \int_0^\infty \frac{\cos(\omega\tau)}{\omega^3} d\omega \tag{54}$$

Since we integrate only over the positive half-line, the absolute value signs in (53) - (54) can be omitted.

However, in (53) and (54) the integrand diverges as $\omega \to 0$. This singularity of the spectral density at low frequency (and the resulting divergence of (54)) indicates that the random function belongs to the class of Intrinsic Random Functions (IRF) of order 1 (Matheron [12]). In particular, the variance of the random function is infinite, but its first derivatives (or, equivalently, increments along arbitrary directions) form a stationary random field. The function $k_f(\tau)$ obtained below in (64) after regularization is a generalized covariance for this IRF(1) class. Introducing a small parameter $\omega_0 > 0$ provides one form of regularization, equivalent to imposing constraints on the low-frequency behavior of the function space, and yields a well-defined covariance suitable for practical computation.

Take a very small $\omega_0 > 0$ and consider a random function analogous to (53), but with all frequencies below ω_0 removed for every realization. Its expression differs from (53) only by the lower limit of integration:

$$F(x) = \int_{\omega_0}^\infty \frac{V_R(\omega)\cos(\omega x) - V_I(\omega)\sin(\omega x)}{\omega\sqrt{\omega d\omega}} d\omega \tag{55}$$

Its correlation function is

$$k_f(\tau) = \int_{\omega_0}^\infty \frac{\cos(\omega\tau)}{\omega^3} d\omega \tag{56}$$

A harmonic oscillation of frequency $\omega_0$ has period

$$T_0 = \frac{2\pi}{\omega_0} \tag{57}$$

Choose $\omega_0$ so that $T_0$ is much larger than the range of x-values in the training sample. Then, within that domain, all oscillations with frequencies below $\omega_0$ degenerate into approximately linear terms and differ little from one another (or from frequencies slightly above $\omega_0$).

Therefore, if the inputs $x_i$ lie in a bounded interval, one can always choose $\omega_0$ sufficiently small that (55) - (56) serve as a practically acceptable replacement for (53) - (54); as $\omega_0 \to 0$, the two become equivalent.

Integrate (56) by parts twice:

$$k_f(\tau) = \int_{\omega_0}^{\infty} \frac{\cos(\omega\tau)}{\omega^3} d\omega = -\frac{\cos(\omega\tau)}{2\omega^2}\bigg|_{\omega_0}^{\infty} - \tau \int_{\omega_0}^{\infty} \frac{\sin(\omega\tau)}{2\omega^2} d\omega =$$

$$= \left(-\frac{\cos(\omega\tau)}{2\omega^2} + \tau\frac{\sin(\omega\tau)}{2\omega}\right)\bigg|_{\omega_0}^{\infty} - \frac{\tau^2}{2}\int_{\omega_0}^{\infty} \frac{\cos(\omega\tau)}{\omega} d\omega \tag{58}$$

The last term involves the cosine integral. Considering $\tau \geq 0$,

$$-\int_{\omega_0}^{\infty} \frac{\cos(\omega\tau)}{\omega} d\omega = -\int_{\omega_0\tau}^{\infty} \frac{\cos(\omega\tau)}{\omega\tau} d(\omega\tau) =$$

$$= \gamma + \ln(\omega_0\tau) + \int_0^{\omega_0\tau} \frac{\cos(\omega) - 1}{\omega} d\omega\,, \tag{59}$$

where $\gamma$ is the Euler-Mascheroni constant.

Substituting this into (58) gives

$$\frac{\cos(\omega_0\tau)}{2\omega_0^2} - \tau^2\frac{\sin(\omega_0\tau)}{2\omega_0\tau} + \frac{1}{2}\tau^2\left(\gamma + \ln(\tau) + \ln(\omega_0) + \int_0^{\omega_0\tau} \frac{\cos(\omega) - 1}{\omega} d\omega\right) =$$

$$= \frac{1}{2}\tau^2\left(\ln(\tau) + \left(\ln(\omega_0) + \gamma + \int_0^{\omega_0\tau} \frac{\cos(\omega) - 1}{\omega} d\omega - \frac{\sin(\omega_0\tau)}{2\omega_0\tau}\right)\right) +$$

$$+ \frac{\cos(\omega_0\tau)}{2\omega_0^2} \tag{60}$$

Since, as noted earlier, we may rescale $k_f(\tau)$ by an arbitrary constant (only the ratio $k_f(0)$ to $\sigma^2$ matters), we can drop the factor 1/2 in (60) and write

$$k_f(\tau) = \tau^2\big(\ln(\tau) - b(\tau)\big) + c(\tau), \tag{61}$$

where

$$b(\tau) = \frac{\sin(\omega_0\tau)}{\omega_0\tau} - \ln(\omega_0) - \gamma - \int_0^{\omega_0\tau} \frac{\cos(\omega) - 1}{\omega} d\omega \quad (62)$$

$$c(\tau) = \frac{\cos(\omega_0\tau)}{\omega_0^2} \quad (63)$$

As $\omega_0 \to +0$, both $b(\tau)$ and $c(\tau)$ diverge:

$$\lim_{\omega_0 \to +0} b(\tau) \to +\infty,$$

$$\lim_{\omega_0 \to +0} c(\tau) \to +\infty$$

And the random function described by (55) - (56) tends to that described by (53) - (54). Note that $c(\tau)$ diverges much faster than $b(\tau)$, whose growth is dominated by $-\ln(\omega_0)$.

Thus we obtain the regularized covariance function (61) for the one-dimensional case and $\tau \in [0, +\infty)$. It is straightforward to extend it to higher dimensions, since the covariance must be radially symmetric. Moreover, if $\omega_0$ is chosen sufficiently small so that the period (57) is much larger than the range of τ of interest, then $b(\tau)$ and $c(\tau)$ are effectively constants. Making this replacement in (61) yields a generalized covariance that is convenient for computation.

Final result (generalized covariance):

$$k_f(\tau) = \|\tau\|^2(\ln(\|\tau\|) - b) + c\,, \tau \in R^n \quad (64)$$

where b and c are constants that can be estimated using (62) and (63).

The function (64) is a generalized covariance for the IRF(1) class [12]. In what follows we use this term. The constants b and c promote stability by absorbing polynomial components.

Let us restate formulas (29) and (32).

The solution to the approximation problem (1) is

$$f(x) = \sum_{i=1}^{k} \lambda_i k_f(x_i - x) \quad (65)$$

with coefficients determined by

$$\lambda = (K + \sigma^2 E)^{-1} Y \quad (66)$$

where $K$ is the $(k \times k)$ matrix with entries $k_{ij} = k_f(x_i - x_j)$,

$E$ is the identity matrix,

$\lambda$ – column vector $(\lambda_1, \lambda_2, \ldots, \lambda_k)$,

$Y$ – column vector $(y_1, y_2, \ldots, y_k)$,

$\sigma^2$ is the variance of the random variables $u_i$ in (1).

The linear combination (65) with kernel (64) is also known as a polyharmonic spline (notably, the thin plate spline) (R. L. Harder, R. N. Desmarais; F.L. Bookstein [10], [4]). In the present setting, there are some practical nuances: the constants b and c are estimated via (62) - (63), and the variance $\sigma^2$ of the $u_i$ is introduced directly into the linear system (66) to solve an approximation (rather than exact interpolation) problem.

Moreover, the function (64) corresponds to the generalized polyharmonic spline in $R^n$ (Buhmann; Wendland [5], [18]). In particular, the term $\tau^2 \ln\|\tau\|$ is the canonical kernel for the minimum-penalty spline that is invariant under translations, rotations, and scalings in $R^n$. Importantly, in this work the kernel's form is not chosen heuristically (nor by analogy with the 2D case) but is derived directly from the symmetry postulates 1-3 of the probability measure in the infinite-dimensional function space, and it is valid for arbitrary input dimension n.

The generalized covariance (64) is also closely related to Duchon splines (Duchon, 1977 [8]) with smoothness order m = 2. Its leading term $\tau^2 \ln\|\tau\|$ matches exactly the canonical Duchon kernel in dimension n = 2. For other dimensions, the structure of (64), derived from symmetry and regularization, ensures stability and includes the components $-b\|\tau\|^2 + c$ corresponding to the polynomial null space of the operator – standard when working with intrinsic random functions (Matheron, 1973 [12]). Thus, the result provides a probabilistic interpretation and justification for the class of smoothing methods based on Duchon splines.

As for the linear system (66) (equivalently (31)), it is mathematically identical to Gaussian Process Regression (GPR) (Rasmussen & Williams, 2006 [14]) with covariance $k_f$ and noise variance $\sigma^2$. In broader machine-learning terms, it is also equivalent to Kernel Ridge Regression (KRR) (Saunders et al., 1998; Bishop, 2006 [15], [2]) in the reproducing kernel Hilbert space (RKHS) induced by $k_f$, where $\sigma^2$ plays the role of the L2 regularization parameter.

The key result and novelty of this work lie not in the form of the solution equations, but in the rigorous derivation of the explicit kernel (64) from first principles - namely, the symmetry postulates 1-3 for the probability measure in an infinite-dimensional function space.

Let us now discuss properties of the matrix K in (66). It is built from the generalized covariance (64) and may be singular due to the presence of polynomial components (constants and linear terms) in its null space (Chilès & Delfiner, 2012, ch. 4 [6]). In practice, this can occur if distances between sample points satisfy certain relations (e.g., for two points when $\|x_1 - x_2\| = e^b$). However, the following factors typically prevent issues:

The parameter $\sigma^2 > 0$ in (66) guarantees that $K + \sigma^2 E$ is invertible.

Typical values of b (determined via $\omega_0$ in (62)) make the critical distance $e^b$ extremely large (for example, with $\omega_0 = 0.001$, b ≈ 7.33, so $e^b$ ≈ 1500), far exceeding characteristic inter-point distances within a bounded approximation domain.

For exact interpolation problems ($\sigma^2 = 0$), even under unfavorable point configurations one can use the function (61), computing b and c via (62) - (63), and also apply the standard polyharmonic spline procedure with an explicit polynomial part and orthogonality conditions (Wahba, 1990 [17]).

Example.

As an illustrative example, consider the one-dimensional approximation setting. Take a training set whose inputs mostly lie in the interval [0, 10]; thus, for estimating b and c, assume $\tau$ ranges from 0 to 10.

Choose $\omega_0 = 0.001$. The corresponding period (57) is $T_0 = 6283.1853$, i.e., roughly 600 times larger than typical differences between the $x_i$ values. It is reasonable to regard frequencies below $\omega_0$ (with even longer periods) as negligible for solving the task on our approximation domain.

Compare $b(\tau)$ and $c(\tau)$ from (62) - (63) at $\omega_0 = 0.001$ for $\tau = 0$ and $\tau = 10$:

$$\mathrm{b}(0) = 1 + 6.907755 - 0.577216 + 0 = 7.33054$$

$$b(10) = 0.999983 + 6.907755 - 0.577216 + 0.000025 = 7.330548$$

$$c(0) = 1000000$$

$$c(10) = 999950.000416$$

The values at τ = 10 differ from those at τ = 0 by only 0.0001% (for b) and 0.005% (for c). We therefore take them as constants: b = 7.33054 and c = 1,000,000.

Let us first set $\sigma^2$ = 0 in (66). In this case all $u_i$ in (1) are zero, and $f(x)$ must pass exactly through all training points; the approximation problem becomes an interpolation problem. This behavior is observed in Figure 1.

Note. When computing entries of $K$ via (64), recall that the expression $\|\tau\|^2 \ln(\|\tau\|)$ is undefined at $\|\tau\| = 0$. However, since

$$\lim_{\|\tau\|\to 0} \|\tau\|^2 \ln(\|\tau\|) = 0,$$

for diagonal entries of $K$ (where $x_i = x_j$ and $\|\tau\| = 0$) one should explicitly set

$$k_f(0) = c.$$

For off-diagonal entries with extremely small separations $\|\tau\|$ (e.g., $\ll 10^{-10}$) it is also advisable to use the approximation

$$k_f(\tau) \approx c$$

to avoid numerical instabilities.

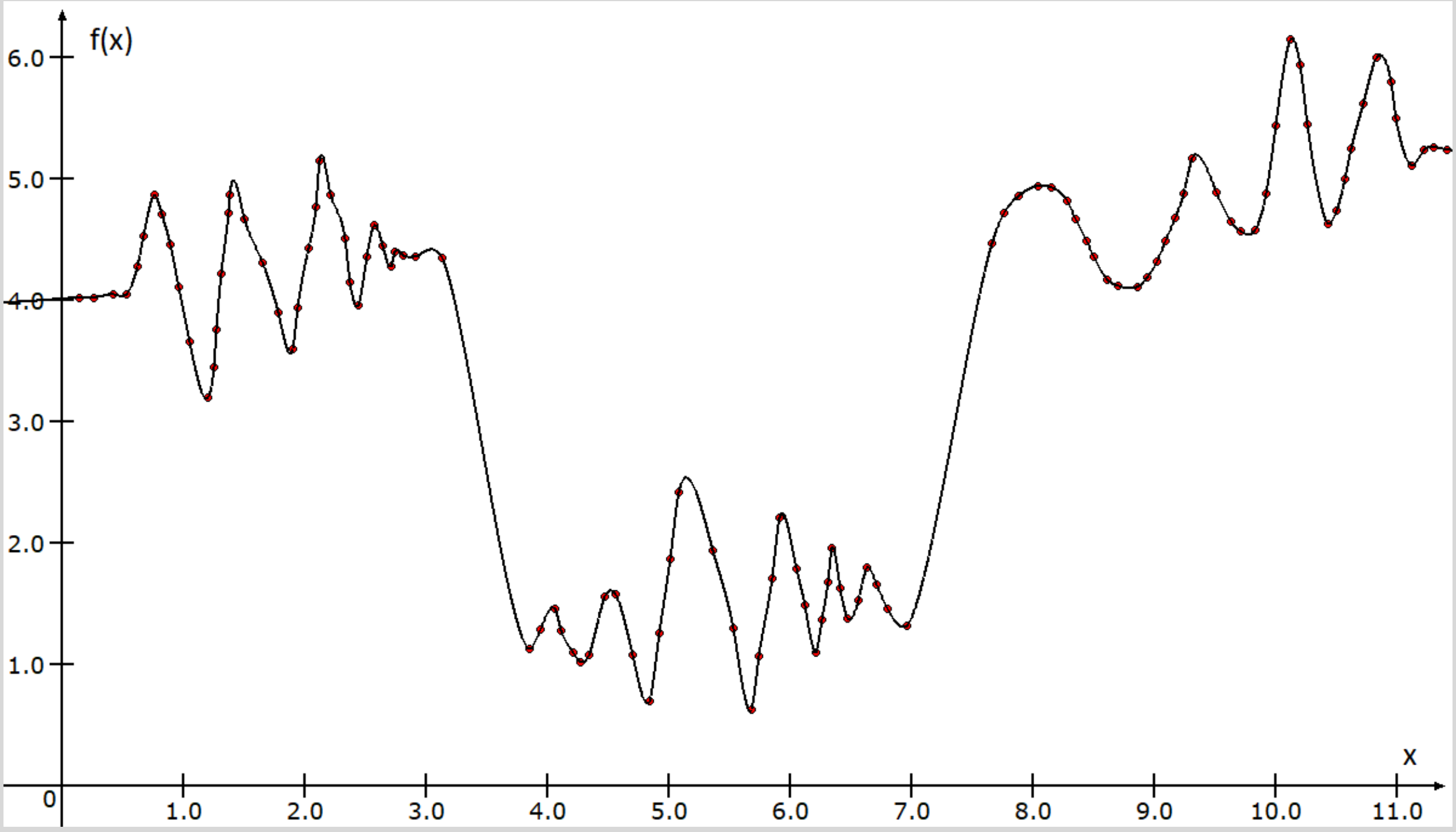

Figure 1.

The function (65) readily reproduces complex nonlinear dependencies without jumps or oscillations between adjacent points.

If we take $\sigma^2 > 0$, we return to the original approximation setting (1).

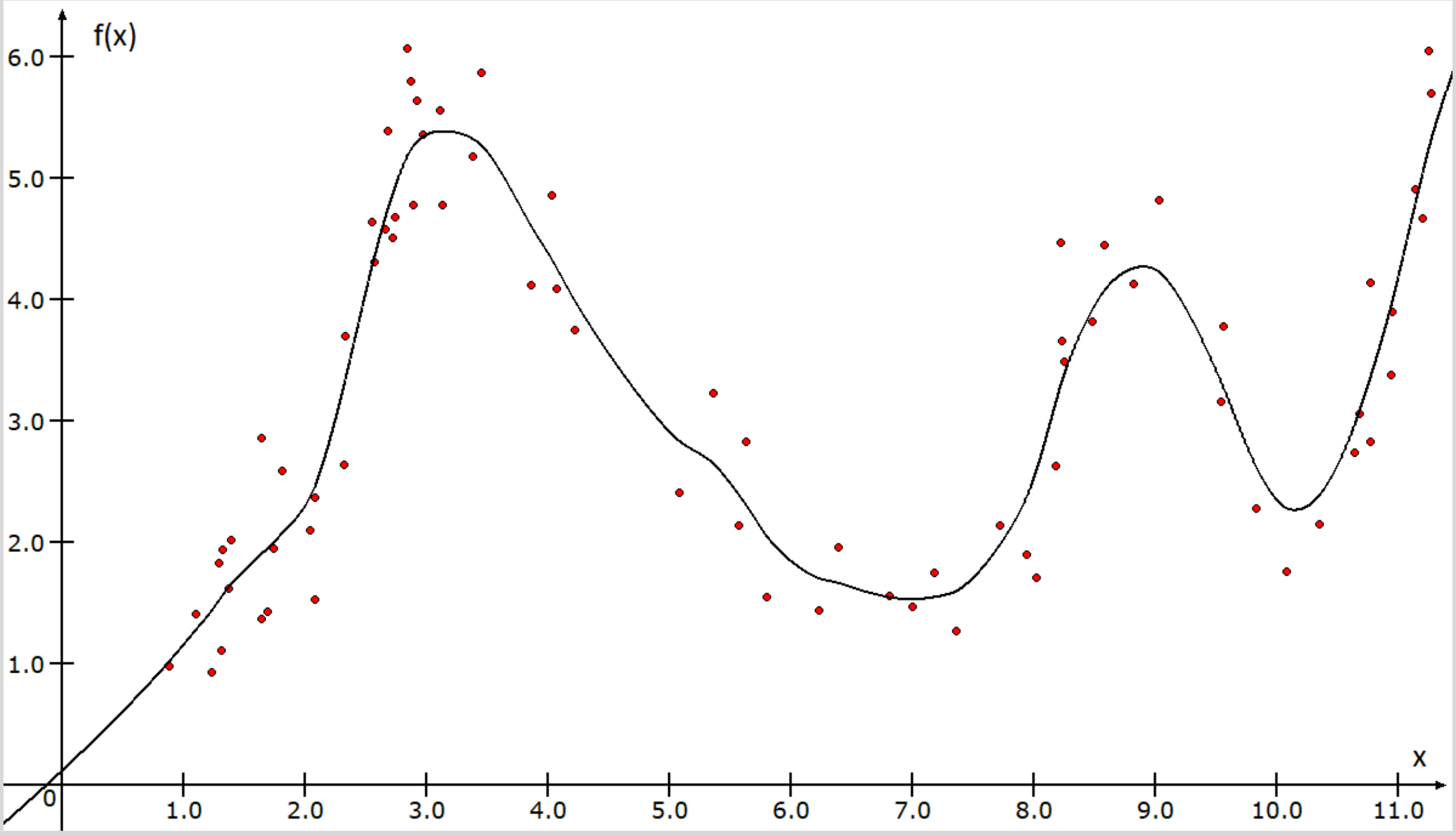

Figure 2.

In the example shown in Figure 2, $\sigma^2$ = 0.625.

**Discussion.**

The solution (65), (66) with kernel (64) enjoys several fundamental advantages stemming from its spectral properties and principled derivation:

1. Stability and absence of oscillations. Unlike many other kernels (e.g., Gaussian), the corresponding spectral density (45) has a power-law form $S(\omega) = a\|\omega\|^{-(n+2)}$. Such a spectrum has no "holes" nor exponentially fast decay at high frequencies. Consequently, the basis induced by this kernel is "complete" over a broad function class and is not prone to oscillations between interpolation nodes - a frequent issue with, for example, Gaussian radial basis functions, which can lead to overfitting and instability (Gibbs phenomenon). The resulting surfaces are smooth and predictable.
2. Natural regularization via $\sigma^2$. The parameter $\sigma^2$ provides a smooth transition from exact interpolation ($\sigma^2 = 0$) to approximation ($\sigma^2 > 0$), effectively combating overfitting. Unlike many methods, even in the interpolation regime the solution remains stable thanks to the kernel's properties.
3. Principled kernel choice. Typically, the kernel (e.g., Gaussian, polynomial) in methods such as Kernel Ridge Regression or Gaussian Process Regression is chosen empirically, often by trial and error. Here, the kernel is not chosen but derived from fundamental indifference and symmetry principles. Thus, when no prior information is available, the polyharmonic spline (64) is not merely an option – it is the natural a priori choice. Using a kernel with a different spectrum (e.g., rapidly decaying as with the Gaussian) amounts to an unwarranted prior assumption that high-frequency components in the target $f(x)$ are unlikely, which may either lose detail or induce oscillations when attempting to recover it.

Therefore, the proposed method not only provides concrete solution formulas but also offers a theoretical guarantee of their stability and adequacy for a wide range of approximation problems in the complete absence of prior information.

**Conclusion.**

This paper has shown that a machine learning regression problem, formulated as a multivariate approximation task in (1), can be solved within the theory of random functions by assuming the existence of a probability measure on the function space that satisfies the symmetry postulates of items 1-3. The solution is a linear combination of the correlation function, whose spectral density (45) - (46) can itself be derived from the scaling symmetry (4). It was further shown that the resulting generalized covariance corresponds to a polyharmonic spline of the form (64). In summary, formulas (62) - (66) provide the solution to (1).

Unlike most approaches in machine learning, where the kernel or model structure is chosen empirically, here the entire solution architecture – including the form of the basis functions, the type of regularization, and the noise parameterization – is derived analytically from a minimal set of natural symmetry assumptions about the function space. This allows one to view the polyharmonic spline not merely as a convenient approximator, but as a theoretically justified solution in the complete absence of prior information.

We note in closing that Wieler (2022) [19] independently arrived at conclusions consistent with those of this paper, showing that the polyharmonic spline emerges as a solution to the regression problem under minimal assumptions: invariance under translation, scaling, and rotation, together with a prescribed degree of smoothness. However, unlike the present work – which uses a Gaussian prior measure and the apparatus of random function theory – Wieler constructs a non-Gaussian scale-invariant process (SIP), leading to a t-process and a form of natural regularization that does not require hyperparameter tuning. The fact that two fundamentally different approaches – one rooted in classical random function theory, the other in the spirit of objective Bayesian inference-converge to the same kernel underscores the fundamental role of the polyharmonic spline as the “natural” choice in the absence of prior information.

It is important to emphasize that the Gaussianity postulate for the probability measure on the function space (item 3) is not introduced to impose any characteristic amplitude scale, but to guarantee the existence of the integral canonical representation (8) – a key tool in the theory of random functions. Without this step, one cannot rigorously pass from an abstract measure to the spectral representation (10). Moreover, since no specific variance is fixed, Gaussianity does not introduce a scale and remains fully consistent with the indifference principle. In this way, Gaussianity enables the use of the random function framework, while the kernel's form is determined entirely by the symmetries of the space.

**Acknowledgments.**

The author thanks the anonymous reviewers for valuable feedback on earlier versions of this work.